\documentclass{ExcelAtFIT}

\ExcelFinalCopy

\hypersetup{
	pdftitle={Generative Adversarial Networks Applied for Privacy Preservation in Biometric-Based Authentication and Identification},
	pdfauthor={Lubos Mjachky and Ivan Homoliak},
	pdfkeywords={Privacy Preservation, Machine Learning, Generative Adversarial Networks}
}

\ExcelYear{2021}

\PaperTitle{Generative Adversarial Networks Applied for Privacy Preservation in Biometric-Based Authentication and Identification}

\Authors{Lubos Mjachky and Ivan Homoliak}
\affiliation{%
  \href{mailto:homoliak@fit.vutbr.cz}{homoliak@fit.vutbr.cz},
  \textit{Brno University of Technology, Faculty of Information Technology}}


\Supplementary{\href{https://github.com/lubosmj/I2I-GANs}{Downloadable Code}}

\Abstract{
Biometric-based authentication systems are getting broadly adopted in many areas. However, these systems do not allow participating users to influence the way their data are used. Furthermore, the data may leak and can be misused without the users’ knowledge. In this paper, we propose a~new authentication method that preserves the privacy of individuals and is based on a generative adversarial network (GAN). Concretely, we suggest using the GAN for translating images of faces to a visually private domain (e.g., flowers or shoes). Classifiers, which are used for authentication purposes, are then trained on the images from the visually private domain. Based on our experiments, the method is robust against attacks and still provides meaningful utility.
}

\Teaser{
	\TeaserImage{gan-system-teaser.pdf}
}

\begin{document}

\startdocument


\section{Introduction}

Biometric-based authentication systems are being used by millions of people daily. Every modern smartphone is equipped with facial recognition hardware. However, information about faces may end up in an unreliable place and users cannot influence that.

This opens a question of trust because the privacy of an individual should be always protected unless there is a public interest in revealing it. Most of the time, it is sufficient to lean towards obfuscating images by blurring, masking, applying pixel-level modifications \cite{video-anonymize-faces}, or introducing additional noise \cite{ganobfuscator, dpgan-zhang}. But, these techniques do not provide much utility.

Several researchers started to advocate the usage of more robust techniques. Chen et al. \cite{gan-based-privacy-preservation-springl-chen} and Sirichotedumrong and Kiya \cite{gan-based-transformation-scheme-siri} used a generative adversarial network (GAN) to translate input images into a visually protected domain. Similarly, Ito et al. \cite{better-scheme-siri-ito} proposed a transformation neural network that is trained in a way so that the generated images reduce the loss value of a final classification model. All these mechanisms closely relate to the method proposed in this paper. However, none of them considers a~scenario where both the server and clients are mutually distrustful and the communication between them is not curated by a~trusted third party.

In this paper, we aim our attention to defining a~novel method for preserving privacy by utilizing a~GAN. The GAN is used to translate images of faces to a visually private domain (e.g., flowers or shoes), totally unrelated to the domain of faces. Thus, the privacy is assured solely by the GAN and the learnt mapping function between the domains.

\paragraph{Contributions.}
Overall, our contributions are twofold. First, we proposed a method that does not require any extra transformation to perform a translation to a~visually private domain, as opposed to the previous works. Second, we carried out experiments on multiple target domains and successfully validated the proposed method on real-world binary classification tasks, representing a centralized authentication use case.

\section{Generative Adversarial Networks}

A Generative Adversarial Network (GAN) consists of two neural network models: a \textit{generator} and a \textit{discriminator}. The generative model generates fake samples and the discriminative model strives to determine whether a sample comes from the real data distribution or the generative model distribution. Both of these models are trained simultaneously and they play the two-player minimax game.

During the training, the discriminator is fed with real and generated samples. The output of the discriminator is recorded during the backpropagation and the weights of the discriminator and generator are then updated correspondingly \cite{goodfellow2014generative}.


GANs can be used for image-to-image translation tasks where the goal is to learn the mapping between two or more distinct domains. The mapping function represents the generative model within the framework. Once there is discovered such mapping, we can transfer styles or textures from one domain to another.

\subsection{Common Architectures}
\label{ssec:common-gans}

The standard GAN objective is not robust enough. For many trainings, the generative models may fail to converge or lead to mode collapse. To deal with this problem, additional losses have to be introduced to the training schema. For example, CycleGAN \cite{zhu2017cyclegan}, DualGAN \cite{yi2017dualgan}, and DiscoGAN \cite{kim2017discogan} use cycle-consistency or reconstruction loss. The authors of these frameworks opted for introducing two mapping functions, i.e., for translating images from a first domain to a~second domain, and vice versa. These two mapping functions (generators) together aspire to minimize the cycle-consistency or reconstruction loss.

Models using the loss function considering only cycle-consistency can unnecessarily prefer an easily invertible mapping function. Such behaviour is not appropriate when the mapping function should be rather complex. Furthermore, generators must learn invertible mappings which need to be each other's inverses. Due to this limitation, some frameworks suggest bringing back the idea of having a single generator and thus circumventing cycle-consistency constraints or their derivations. GcGAN \cite{fu2019gcgan} attempts to approximate just one mapping function by employing geometry-consistency loss. TraVeLGAN \cite{amodio2019travelgan} utilizes a third siamese network in addition to the discriminator and generator to capture high-level semantics between translated domains.

On the other hand, combining cycle-consistency loss with other losses adds benefits to the training process significantly. SPA-GAN \cite{emami2020spa} trains the model by leveraging the standard GAN loss, cycle-consistency loss, and feature map loss. Here, a so-called attention mechanism embedded directly to the GAN architecture is utilized to allow the model to focus more on the most distinctive regions in images. U-GAT-IT \cite{kim2020ugatit} is a~GAN framework that uses an improved version of the standard GAN loss (the loss introduced in LSGAN \cite{mao2017lsgan}), cycle-consistency loss, identity loss, and class attention maps loss. In style transfer tasks, U-GAT-IT outperforms other GAN frameworks by a large margin.

\section{Privacy Aspects of Machine Learning}

All the frameworks introduced in the preceding section operate in a setting where a single model is trained on a centralized dataset.

\subsection{Centralized Learning}

Centralized learning refers to a usual way of machine learning where a neural network has access to a whole dataset. But, a collection of photos, speech, or videos gathered from multiple individuals poses tremendous privacy risks. The users from whom the data were collected can neither control how the data will be used nor delete them. Moreover, researchers are often allowed to perform deep learning only on datasets belonging to their institutions. This can result in an overfitted model that has reduced utility on other inputs \cite{shokri2015privacy}.

Some of the issues can be resolved via differential privacy (DP). DP ensures that when an item is added or removed from a database, it does not affect the outcome of a query \cite{dwork2008differential}. With GANs, it is possible to synthesize private datasets before dispatching them to an insecure environment by introducing noise to the learning procedure. GANobfuscator \cite{ganobfuscator} is a differentially private GAN that adds carefully designed noise to the gradients during the training. PPGAN \cite{zhang2019differentially} directly perturbs the objective function instead. These techniques can be used for anonymizing input datasets.

On the contrary, homomorphic encryption (HE) offers better privacy. HE is a~form of encryption that supports computations on encrypted data without decrypting them first. Therefore, it is not required to possess a secret key to execute computations on the receiver’s side. However, HE calls for extensive refactorization of existing systems and the high computing complexity is still an open problem even though the efficiency of HE schemes is continuously improving. In comparison, employing HE in machine learning is much more complicated than using DP.

\subsection{Collaborative Learning}

There is also an option to incline towards methods that do not require users to share their datasets. Such techniques take advantage of collaborative learning (a.k.a., federated learning), where training data does not leave users’ devices.

Shokri and Shmatikov \cite{shokri2015privacy} designed a~system that allows multiple parties to jointly learn local neural network models. They exploited the fact that stochastic gradient descent (SGD) can be parallelized and executed asynchronously. The system works as follows. Enrolled parties train their local models concurrently and independently while selectively sharing some of the model parameters, namely gradients. The parameter sharing approach enables the parties to benefit from each other because the parameters obtained from different users avoid the local models being stuck in local minima. In this system, DP is used to preserve the privacy of individuals by applying noise to the parameters sent by the participants.

Although DP provides protection, it is crucial to use it properly. Hitaj et al. \cite{hitaj2017deep} investigated the system proposed by Shokri and Shmatikov and found out that the level of granularity was not defined correctly for DP. They successfully performed an attack that consisted of an active adversary which could influence the learning process. Because of that, the genuine participants were leaking data from their private datasets during the training.

Setting user or device-level DP should be efficient against the active attack devised earlier according to Lim et al. \cite{lim2020federated}. Lim et al. also noted that it is necessary to have a curator that aggregates and randomly selects a~group of participants who train a model for each iteration. In this case, a malicious participant should not be able to extract information about other participants since it is unclear who has participated in a single training round. Yet, the authors of the paper \cite{wang2019beyond} were able to recover data even when the learning procedure used the user-level DP.

\section{Proposed Method}
\label{sec:proposed-method}

Despite recent progress, collaborative learning still faces many challenges. A serious bottleneck is communication. Because of that, it is required to implement communication-efficient protocols that iteratively send small amounts of data through the network. Other issues relate to the system and statistical heterogeneity. Every participant within the system can have different computational capabilities and data points across devices may vary significantly \cite{li2020federated}.

Centralized learning provides fewer opportunities for adversaries in comparison to collaborative learning. But, at the cost of privacy. To tackle the privacy problem, we propose replacing sensitive data from a~private domain with data from a safe domain. In particular, the translation to the safe domain can be performed by a GAN. Then, the centralized model is trained purely on the outputs of the GAN and is therefore unaware of sensitive data points from the original distribution.

\begin{figure}[t]
    \centering
    \includegraphics[width=\linewidth]{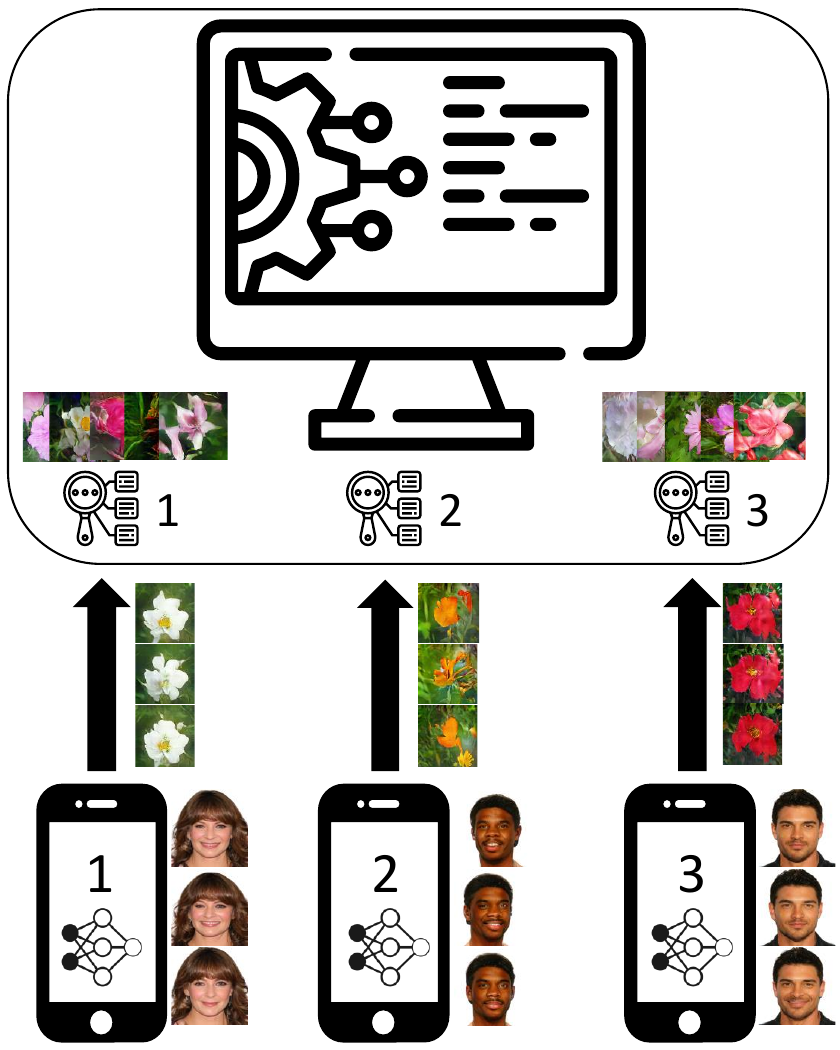}
    \captionof{figure}{A GAN-based authentication system. Face images of users are translated to images of flowers which are then used for authentication purposes.}
    \label{fig:proposed-auth-architecture}
\end{figure}

This means that we can use the translated images for authentication purposes. For instance, we assume scenarios in which users authenticate themselves with images of flowers or shoes without revealing their actual identities, like depicted in Figure \ref{fig:proposed-auth-architecture}. As opposed to standard biometric-based systems, here, the classifiers are learnt how to identify users concerning the features posed by images of flowers or shoes instead of real faces.

By combining centralized learning and the presented authentication method, we expect that the privacy of individuals should be protected against active and passive adversaries. Passive adversaries that captured synthetic images should not be able to reconstruct face images that resemble real users. Equivalently, active adversaries should not be able to create a~reverse mapping from a visually private domain back to a domain of faces by harassing the learning process. Eventually, such active attacks often lead to model poisoning which can be detected on both the server and client sides \cite{zhao2020shielding}.

The privacy protection is provided by a GAN which resides on the user's device and is trained on arbitrary datasets. In general, users will likely train their GAN models on a private dataset backed by public datasets. Once an adversary gains access to the user's device, GAN, and datasets, the privacy may be violated. The adversary can then insert a new inverse GAN into the training framework, freeze the layers of the user's GAN, and start learning the reverse mapping. In the end, the adversary has a generator capable of generating the actual faces of the user.

\section{Performed Experiments}

To validate the proposed method, we decided to perform multiple experiments. First, we conducted a study of appropriate datasets required for training GANs. Then, we evaluated the performance of binary classifiers on synthetic images produced by the GANs. And finally, we devised a plan for attacking our privacy-protecting procedure.

\subsection{Datasets}

Assuming that the input data for the translation task are always images of faces, one of the best publicly available datasets is the CelebA dataset \cite{liu2015celeba}. The images cover large pose variations, different skin tones, and face shapes. Such diversity is a prerequisite for good generalization properties.

Choosing a satisfying output domain was tricky. We have determined that many datasets are very heterogeneous and asymmetric to the domain of faces. Ultimately, the following datasets were considered for further evaluation: shoes \cite{yu2017semanticzappos}, textures \cite{cimpoi2014dtd}, cars \cite{stanford2013cars}, flowers \cite{nilsback2008flowers}, and food \cite{bossard14food}.

All these datasets, except for one, provide sufficient variability across the classes. In the dataset of shoes, further augmentations need to be employed since the images are captured just from one angle.

\subsection{Used Architectures}

There are plenty of GAN frameworks suitable for translating images from one domain to another. We have selected only four of them for our intention: CycleGAN, DiscoGAN, TraVeLGAN, U-GAT-IT.

In the experiments, CycleGAN performed worst. When trained on the datasets of shoes and textures, the loss values of the discriminators rapidly converged to zero causing immediate mode collapse after no more than 20 epochs (half day) of training. We suppose that this behaviour is linked to the disadvantages of cycle-consistency loss, as we stated in Section \ref{ssec:common-gans}. On the other hand, proper hyper-parameters tuning, like setting different weights for the loss functions, could resolve such a problem. Yet, we found this hyper-parameters tuning time consuming and disregarded CycleGAN for additional experiments.

The DiscoGAN framework worked better on all datasets. But, even after introducing feature matching loss \cite{salimans2016improved} to the training process, the model could not learn to generate sharp and crisp images. The inability to generate plausible images led us to experiment with the loss established in the WGAN-GP paper \cite{gulrajani2017improved}. Also, we tried to use spectral normalization \cite{miyato2018spectral}. Overall, these adjustments did not improve the final results. The generated images looked best only when the model was trained on the dataset of shoes.

\begin{figure}[ht]
    \centering
    \begin{flushleft}
    {\tiny \hspace{1.3cm}
    \textbf{Input} \hspace{1.05cm} \textbf{Output} \hspace{1.0cm} \textbf{Input} \hspace{0.9cm} \textbf{Output}}
    \end{flushleft}
    \includegraphics[width=0.8\linewidth]{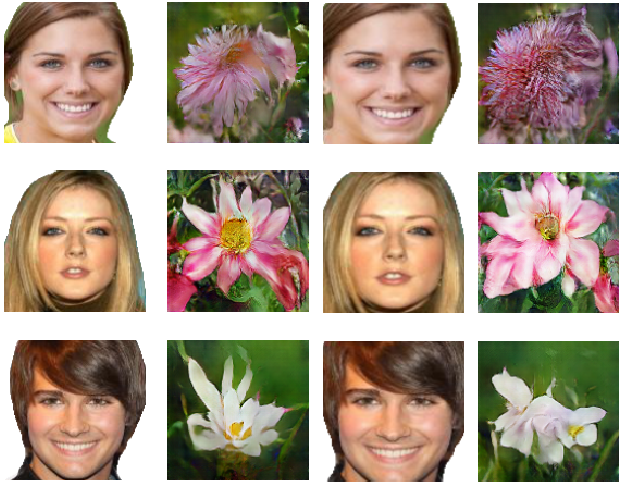}
    \captionof{figure}{Outputs produced by TraVeLGAN. The model was trained on 8,000 cropped CelebA images with removed background. The output domain represents images of flowers.}
    \label{fig:travelgan-resized-white-faces-zoom-samples}
\end{figure}

TraVeLGAN has shown the most satisfactory results out of all tested frameworks. Except for the textures and cars datasets, the model performed well on the rest of the datasets. The models which were trained on the datasets of textures and cars were not producing reasonable outputs and the generative model was jumping between modes for the same input faces through the training. Images generated by TraVeLGAN are demonstrated in Figure \ref{fig:travelgan-resized-white-faces-zoom-samples}. It is important to emphasise that the translations to flowers were only slightly affected by zooming or additional background changes during the experiments.

We also tried to evaluate the performance of U-GAT-IT. In this case, the resulting synthetic images were fairly plausible for the dataset of shoes. However, we were not able to exploit the real performance of U-GAT-IT because the full model requires 32 GB of memory on a single GPU. Furthermore, training the light model for 50 epochs on a dataset that consisted of 10,000 samples took more than 4 days on NVIDIA Tesla T4 GPU.

\subsection{Classification}

According to the experiments, we believe that TraVeLGAN trained on the flowers dataset is a sufficient solution to our problem. The framework correctly identifies the key features of images and translates them to a visually private domain (as shown in Figure \ref{fig:travelgan-resized-white-faces-zoom-samples}).

\begin{figure}[ht]
    \centering
    \begin{flushleft}
    {\tiny \hspace{1.3cm}
    \textbf{Input} \hspace{0.95cm} \textbf{Output} \hspace{1.0cm} \textbf{Input} \hspace{0.9cm} \textbf{Output}}
    \end{flushleft}
    \includegraphics[width=0.8\linewidth]{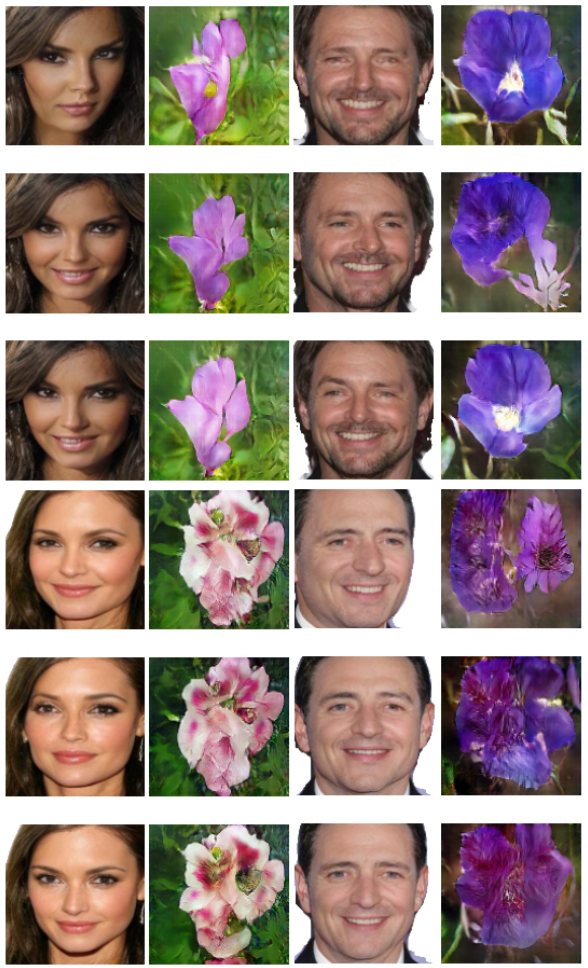}
    \caption{Flowers generated by TraVeLGAN for same identities. Minor head rotations do not influence the outcome of the translation.}
    \label{fig:travelgan-same-identities-augmented}
\end{figure}

To classify the images of flowers, it is necessary to ensure that the model generates undoubtedly similar images for the same individual. Unfortunately, the face images of a single individual contained within the CelebA dataset often do not resemble each other. Because of that, we had to augment the images with StarGAN \cite{choi2020starganv2}. The required precondition for the classification purposes could be afterwards verified (see Figure \ref{fig:travelgan-same-identities-augmented}).

Regarding the augmentation, we noticed that different skin tones contribute to a simple colour change of flowers. On the contrary, facial expressions, as well as hairstyles, affect the translation process significantly, leading to generating flowers of completely different shapes.

We tested the usability of our method on binary classification problems. In the beginning, we selected 93 celebrities from the CelebA dataset and augmented the images of their faces. The number of images ranged from 10 to 20 per selected individual. Next, we translated the augmented images and another random 2,000 images from the CelebA dataset to the images of flowers with TraVeLGAN. Then, we trained 93 binary classifiers.

To evaluate the efficiency of the classifiers, we also studied the performance drop in comparison to standard classification models trained on images of faces. In order to make the conditions for both scenarios similar, we used the same pre-trained MobileNet V2 (MNv2) model\footnote{\url{https://tfhub.dev/google/imagenet/mobilenet_v2_100_128/feature_vector/4}} and adjusted the last output layers for binary classification. Moreover, we performed 5-fold cross-validation and used the following metrics: accuracy, precision, recall, and F1 score.

Table \ref{tab:classification-metrics} shows average scores of the classifiers. The values were averaged over distinct classes (i.e., identities) across the folds. The recall score has revealed that the classifiers trained on the synthetic flowers did not identify almost 30\% out of all positive examples when the layers of MNv2 were frozen. On the other hand, the classifiers correctly identified more than 86\% out of all examples when fine-tuning was enabled.

\begin{table}[h]
	\vskip6pt
	\caption{Actual scores of the evaluated classifiers. The acceptance threshold for the classifiers was set to 0.7. The classifiers were trained for 15 epochs in two settings: (1)~with frozen MNv2 layers, (2)~with trainable MNv2 layers.}
	\centering
	\begin{tabular}{lrrrr}
		\toprule
		& \multicolumn{2}{c}{\textbf{Frozen MNv2}} & \multicolumn{2}{c}{\textbf{Trainable MNv2}} \\
		\cmidrule(r){2-5}
		\textbf{Metric} & \textbf{Faces} & \textbf{Flowers} & \textbf{Faces} & \textbf{Flowers}  \\
		\midrule
		\textbf{Accuracy} & 0.9998 & 0.9976 & 0.9992 & 0.9974 \\
		\textbf{Precision} & 0.9139 & 0.8353 & 0.8810 & 0.8594 \\
		\textbf{Recall} & 0.9148 & 0.7086 & 0.9117 & 0.8627 \\
		\textbf{F1 Score} & 0.9143 & 0.7667 & 0.8961 & 0.8611 \\
		\bottomrule
	\end{tabular}
	\label{tab:classification-metrics}
\end{table}

The experiments ensured us that our method can be still used for authentication since the overall performance drop did not exceed 6\% when the models had the same architecture. Hyper-parameters tuning and further architectural changes should decrease the difference even more.

\subsection{Attack Model}

Suppose attackers determine the used GAN framework, e.g., TraVeLGAN and public datasets. A trivial attack consisting of switching the input and output domains and straightforwardly learning the reverse mapping function is not effective. Based on our observations, the generated faces looked like those depicted in the teaser image.

A more sophisticated attack suggests training the GAN the same way as the victim did. The attackers then freeze the trained model and insert a new model into the framework. The new model is trained to discover a reverse mapping from the visually private domain back to the domain of faces more efficiently due to the presence of a correct pair set. This attack is also referred to as an inverse transformation network attack (ITN-Attack) \cite{better-scheme-siri-ito}.

\begin{figure}[t]
    \centering
    \begin{flushleft}
    {\tiny \hspace{0.9cm}
    \textbf{Original} \hspace{0.7cm} \textbf{Reconstructed} \hspace{0.65cm} \textbf{Original} \hspace{0.75cm} \textbf{Reconstructed}}
    \end{flushleft}
    \includegraphics[width=0.9\linewidth]{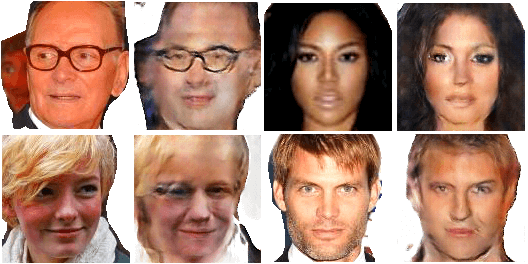}
    \justify
    \captionof{figure}{Images estimated by an inverse transformation network. The network was trained on a correct pair set provided by a TraVelGAN network trained on the datasets of faces and flowers.}
    \label{fig:travelgan-reconstructed}
\end{figure}

We tried to perform an ITN-Attack by embedding a pre-trained TraVeLGAN model into the framework. A couple of reconstructed samples are presented in Figure \ref{fig:travelgan-reconstructed}. From the samples, it is possible to determine sex or hair style, but not the real identity.

\begin{figure}[t]
    \centering
    \begin{flushleft}
    {\tiny \hspace{1.7cm}
    \textbf{Input} \hspace{1.5cm} \textbf{Output} \hspace{1.1cm} \textbf{Reconstructed}}
    \end{flushleft}
    \includegraphics[width=0.8\linewidth]{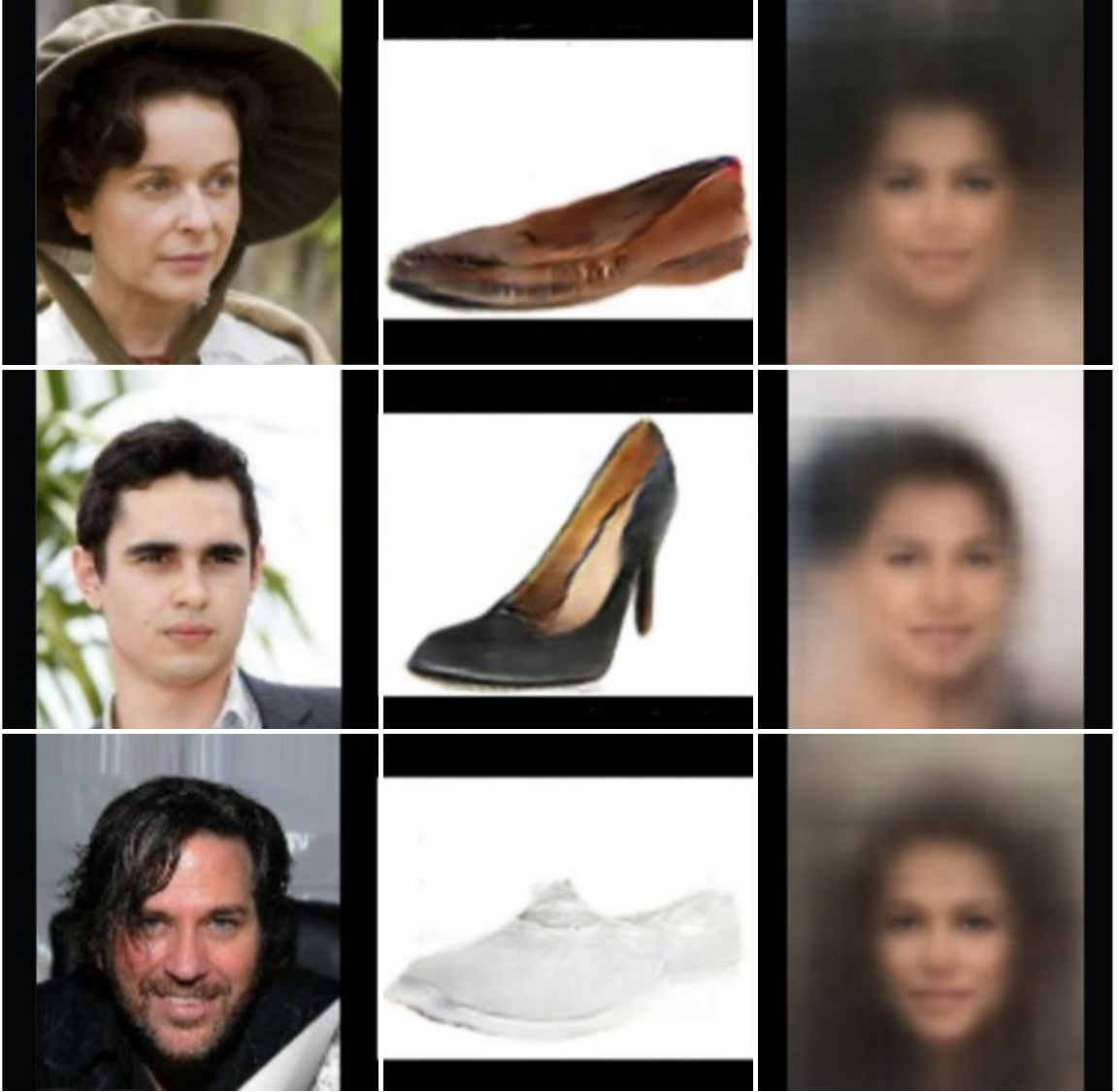}
    \justify
    \captionof{figure}{Outputs of DiscoGAN. The model was trained on 20,000 padded CelebA images. The output domain represents images of shoes. As shown in the last column, DiscoGAN could not properly learn the reverse mapping in a dual setting.}
    \label{fig:discogan-reconstructed}
\end{figure}

Ignoring the fact that the user's models are always initialized with random weights and that the user can govern the hyper-parameters, the attackers may end up having almost identical mapping function from the domain of faces to a visually private domain. Otherwise, there is no guarantee that the attack is going to be successful. Apart from that, the user can utilize DP during the training which results in reduced chances for the attackers as well, yet at the cost of the classification model accuracy. Furthermore, the protection is given by the properties of the chosen datasets. Transforming images between asymmetric datasets back and forth causes losing the information about initial data distribution. We noticed this behaviour when we trained DiscoGAN on the shoes dataset. In a dual setting, the images of faces were accurately translated to the images of shoes, but the opposite did not apply. Mostly, the synthetic images of faces looked unrealistic and blurred (see Figure \ref{fig:discogan-reconstructed}).

\section{Conclusions}

In this paper, we presented a novel approach for employing a GAN for privacy preservation in biometric-based authentication systems. Specifically, the GAN is used to translate face images of individuals to a~visually private domain (e.g., flowers).

The rationale behind the protection of users' privacy lies in the fact that a GAN creates a mapping function that is hard to invert since the target domain is heterogeneous to the domain of faces. Another protection is given by the implicit way of how the GAN can be trained, meaning that different starting point often leads to a different optimal translation. Employing DP with correctly set granularity during the training process naturally increases privacy as shown in many other works \cite{zhu2020moreondp}.

We trained multiple GAN frameworks on different datasets and found out that TraVeLGAN fits best to our problem without any further changes to the architecture. The U-GAT-IT framework has shown to be a good candidate as well, but due to the long training times and high computational requirements, we decided to leave more experiments for future work.

We validated the proposed approach on practical binary classification tasks. The experimental results demonstrated that the images generated by the GAN still provide a~reasonable utility.

To conclude, our method has just two disadvantages: (1) users are required to train GANs on their own which can be time-consuming and (2) some models may end up generating almost identical images for different identities which can result in bad classifiers' performance. On the other hand, the proposed method is not restricted to a specific GAN framework or dataset and does not require training classifiers on the server from scratch since many pre-trained classification networks can be utilized. Therefore, the second disadvantage may be alleviated by using another dataset or a GAN when the problem occurs.

For future work, we will focus on improving the quality of synthetic images as well as the performance of classifiers. We also consider implementing an update procedure allowing users to submit new synthetic images after changing the visage. This could be done by caching the generated images and when the change within a given threshold is detected, the user's device messages the server to consider the new images for authentication.


\phantomsection
\bibliographystyle{unsrt}
\bibliography{2021-ExcelFIT-ShortName-bib}

\end{document}